\documentclass[letterpaper, 10 pt, conference]{ieeeconf}  
\usepackage{hhline}
\usepackage{graphicx}
\usepackage{subcaption}
\usepackage{multirow}
\usepackage{soul,color}
\usepackage{amsfonts}
\usepackage{float}
\usepackage{amsmath}
\usepackage{algorithm}
\usepackage{hhline}
\usepackage{url}
\usepackage{algorithmic}
\usepackage{boldline} 
\usepackage{amssymb} 
\usepackage{pifont} 
\usepackage{booktabs} 
\newcommand{\cmark}{\ding{51}}
\newcommand{\xmark}{\ding{55}}

\usepackage{kotex}
\usepackage{kotex-logo}
\IEEEoverridecommandlockouts                              
\title{\LARGE \bf Multi-task real-robot data with gaze attention for dual-arm fine manipulation}
\author{Heecheol Kim$^{1}$$^{,}$$^{2}$, Yoshiyuki Ohmura$^{1}$, Yasuo Kuniyoshi$^{1}$%
\thanks{$^{1}$ Laboratory for Intelligent Systems and Informatics, Graduate School of Information Science and Technology, The University of Tokyo, 7-3-1 Hongo, Bunkyo-ku, Tokyo, Japan (e-mail: \{h-kim, ohmura, kuniyosh\}@isi.imi.i.u-tokyo.ac.jp, Fax: +81-3-5841-6314) }%
\thanks{$^{2}$ Corresponding author}
\thanks{This study was supported in part by the Department of Social Cooperation Program ``Intelligent Mobility Society Design,'' funded by Toyota Central R\&D Labs., Inc.,  of the Next Generation AI Research Center, The University of Tokyo.}
}
\begin{document}
\maketitle
\thispagestyle{empty}
\pagestyle{empty}
\begin{abstract}
In the field of robotic manipulation, deep imitation learning is recognized as a promising approach for acquiring manipulation skills. Additionally, learning from diverse robot datasets is considered to be a viable method to achieve versatility and adaptability (e.g., \cite{brohan2022rt}). In such research, by learning various tasks, robots achieved generality across multiple objects. However, such multi-task robot datasets have mainly focused on single-arm tasks that are relatively imprecise and not addressed the fine-grained object manipulation that robots are expected to perform in the real world.
In this study, we introduce a dataset for diverse object manipulation that includes dual-arm tasks and/or tasks that require fine manipulation. We generated a dataset containing 224k episodes (150 hours, 1,104 language instructions) that includes dual-arm fine tasks, such as bowl-moving, pencil-case opening, and banana-peeling.  This dataset is publicly available \footnote{The dataset and robot model are available at \url{https://sites.google.com/view/multi-task-fine}}. Additionally, this dataset includes visual attention signals, dual-action labels that separate actions into robust reaching trajectories or precise interactions with objects, and language instructions, all aimed at achieving robust and precise object manipulation. We applied the dataset to our Dual-Action and Attention, which is a model that we designed for fine-grained dual-arm manipulation tasks that is robust to covariate shift. 
We tested the model in over 7k trials for real robot manipulation tasks, which demonstrated its capability to perform fine manipulation.
\end{abstract}
\providecommand{\keywords}[1]{\textbf{\textit{Index terms---}} #1}
 
\begin{keywords}
Imitation Learning,
Deep Learning in Grasping and Manipulation,
Perception for Grasping and Manipulation
\end{keywords}

\section{Introduction}
\begin{figure*}
  \centering
  \begin{subfigure}[t]{0.9\linewidth}
    \centering
    \includegraphics[width=\linewidth]{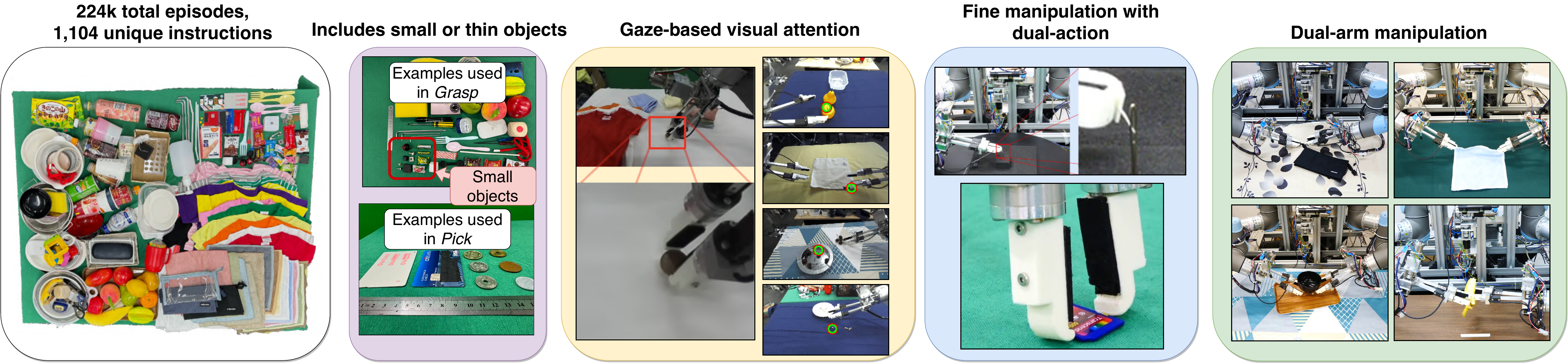}
    \caption{The dataset in this study is a fine-grained robot manipulation dataset that includes dual-arm tasks and a variety of objects, including deformable, small, and thin objects. This dataset includes the gaze of a human demonstrator as a visual attention signal as well as dual-action labels.}
      \label{fig:fig1_concept_dataset}
  \end{subfigure}
  \begin{subfigure}[t]{0.9\linewidth}
    \centering
    \includegraphics[width=\linewidth]{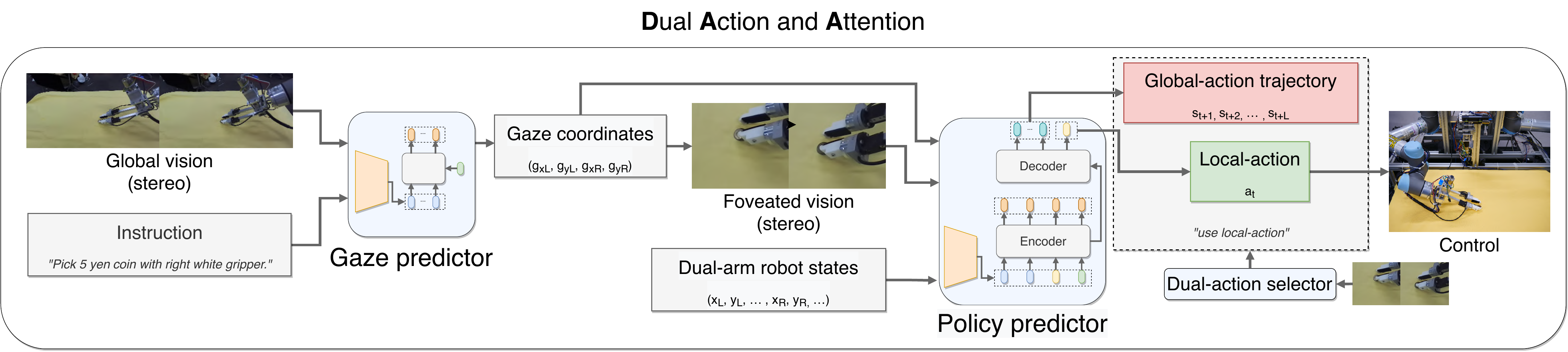}
    \caption{Dual-Action and Attention uses a gaze predictor to efficiently focus on only high-resolution pixels related to the task, thereby enabling manipulation that is robust to task-unrelated objects and background noise. The policy predictor can perform fine-grained manipulation using a dual-action mechanism that simultaneously achieves the generation of a globally robust trajectory (global-action) and precise object manipulation (local-action).}
  \end{subfigure}
  \caption{Dataset and outline of Dual-Action and Attention.}
  \label{fig:fig1_concept}
\end{figure*}
\begin{figure*}
  \centering
  \vspace{0.02in}
  \includegraphics[width=.7\linewidth]{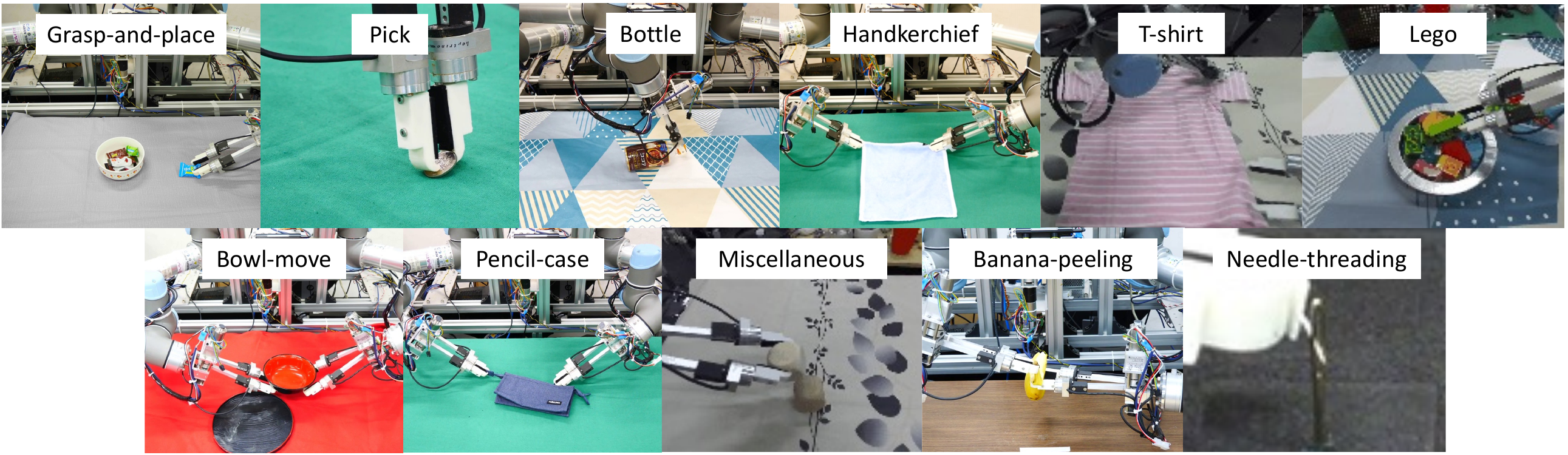}
  \captionsetup{width=.9\linewidth}
  \captionsetup{justification=centering}
  \caption{Visual examples of tasks in the dataset.}
  \label{fig:tasks}
\end{figure*}
 
Fine manipulation, which involves the precise control of delicate objects, is an important research topic because robots are expected to replace fine motor skills traditionally performed by humans in the near future. Deep imitation learning \cite{zhang2018deep,zhao2023learning}, which learns state-action pairs from expert demonstration data via deep neural networks, is a promising approach for fine manipulation because it directly maps a robot's sensory observation to an action, thereby eliminating the need for engineering features and manipulation skills by robot engineers and enabling learning even in cases that involve an unknown environment or object models.

In recent studies on deep imitation learning, researchers have aimed for the generalization of robot manipulation skills by scaling up the dataset size \cite{brohan2022rt, rt22023arxiv,walke2023bridgedata}. These efforts have culminated in the release of \cite{open_x_embodiment_rt_x_2023}, a comprehensive dataset which aggregates over one million episodes from these studies. This large-scale dataset helps efficient research in the robotics community by eliminating the need for individual, time-consuming data generation. However, in many of these studies, researchers primarily focused on the single-arm grasping and placing of objects (e.g., \cite{brohan2022rt,rt22023arxiv}), which lack fine manipulation skills.

We released a robot manipulation dataset containing 224k demonstration episodes that includes fine manipulation and dual-arm skills. This dataset includes data for fine manipulation, such as dual-arm pencil case opening and manipulation of various deformable objects (e.g., thread, banana, and handkerchief), in addition to the manipulation of small and thin objects, such as small bolts, thin coins, and plastic cards. It also includes a human expert's gaze coordinates measured using an eye tracker during demonstrations and dual-action labels to support learning fine manipulation.

To learn fine manipulation, we used and improved the deep learning architecture by implementing Dual-Action and Attention (DAA), which was initially proposed in \cite{kim2021gaze} and later enhanced in \cite{kim2022robot}. In imitation learning, the attention mechanism contributes to a policy inference that is robust to covariate shift -- a scenario in which training and test domain distributions are different \cite{osa2018algorithmic} -- by extracting manipulation-related information from the robot's sensory inputs. \textbf{Visual attention} \cite{kim2020using} is formed by learning human gaze information during demonstrations, creating object-centric high-resolution foveated vision, effectively filtering out task-unrelated features (background and other objects), and enabling the computationally efficient learning of fine manipulation using high-resolution object pixels because this approach excludes task-unrelated pixels from the computation. \textbf{Somatosensory attention} \cite{kim2021transformer} uses a self-attention structure \cite{vaswani2017attention} for robust policy inference, thereby effectively addressing covariate shift in extended robot embodiment, such as dual-arms.

\textbf{Dual-action} was inspired by physiological studies \cite{paillard1996fast} that showed that humans separate approximate reaching and accurate homing in object manipulation. It models global-action for rough object reaching and local-action for precise object manipulation separately. Global-action is robust to compounding errors \cite{ross2011reduction} and local-action allows for precise object manipulation, thus achieving robustness and fine manipulation skills simultaneously. This model enables the learning of fine manipulation tasks, such as needle-threading \cite{kim2021gaze} and banana-peeling \cite{kim2022robot}.

DAA was previously demonstrated in fine manipulation tasks \cite{kim2021gaze,kim2022robot}. In this research, DAA shows that a variety of fine manipulation tasks are possible under various environmental conditions, thereby proving its generality. This model was validated using various tasks, such as opening pencil cases, moving large bowls with both hands, grasping various objects, picking up thin objects, such as coins, and folding towels.

The contributions of this paper are as follows: First, we creat a multi-task imitation learning dataset that includes dual-arm, fine manipulation skills and gaze-based attention information, and make it publicly available. Second, we demonstrate that dual-action and attention mechanisms enable the multi-task dual-arm fine manipulation skills in the dataset to be performed.

\section{Related Work}
\begin{table*}[hbt!]
\centering
\begin{tabular}{lcccl}
\toprule
	\textbf{Dataset} & \textbf{\# episodes} & \textbf{Dual-arm} & \textbf{Visual attention signal} & \textbf{Example skills} \\
\midrule
MIME \cite{sharma2018multiple} & 8.3k & \cmark & \xmark& Grasp, push, stack, pour, open bottles \\
RT-1 \cite{brohan2022rt} & 130k & \xmark & \xmark & Grasp, place, move, knock over, open/close drawer \\
RH20T \cite{fang2023rh20t} & 110k & \xmark& \xmark & Cut, plug, pour, fold \\
BridgeData V2 \cite{walke2023bridgedata} & 60.1k & \xmark & \xmark& Stack, fold clothes, sweep granular materials \\
Mobile Aloha \cite{fu2024mobile} & $<$ 1k & \cmark & \xmark & Push chairs, move bowls, wipe, push buttons \\
DAA (ours) & 224k& \cmark & \cmark & Pick up coins, open zippers, thread needles, peel bananas \\
\bottomrule
\end{tabular}
\caption{The DAA dataset is an at-scale dataset for dual-arm fine manipulation, including signals for learning fine manipulation tasks.}
\label{tab:dataset_summary}
\end{table*}

Imitation learning, particularly behavior cloning \cite{pomerleau1989alvinn}, directly maps a robot's sensory observation to an action in a supervised learning manner. Methods that use deep learning for imitation learning \cite{zhang2018deep,yang2016repeatable} can learn tasks in a model-free manner, which makes them applicable to unknown environments and object dynamics. Deep imitation learning, unlike search-based approaches (e.g., \cite{pinto2016supersizing,openai2019rubiks}) or inverse reinforcement learning (e.g., \cite{das2021model}), which learns a reward function from demonstrations, is safe and sample efficient because it only uses robot demonstrations performed by experts. In studies such as \cite{zhao2023learning,kim2022robot,kim2021gaze,ke2021grasping}, researchers successfully implemented imitation learning that targets fine-grained tasks, such as opening a lid or manipulation with chopsticks. However, in these studies, training was often specialized for one task, which left the expected versatility and adaptability of future robots to new environments/tasks unassessed. Fu \textit{et al.} \cite{fu2024mobile} recently upgraded the robot system originally developed by \cite{zhao2023learning} to a mobile manipulator for dual-arm fine manipulation, such as pushing chairs or cooking; however, their dataset lacks scale (50 demonstrations per task) and does not include visual attention labels, which we consider to be essential for fine manipulation tasks that require high-resolution vision.

Regarding versatility and adaptability, learning multi-task manipulation skills using a large robot dataset is a topic that has been extensively researched recently. In many studies, researchers were encouraged by the advancements in language models \cite{brown2020language} and adopted imitation learning conditioned on natural language instructions as a breakthrough in multi-task learning. 
Sharma \textit{et al.} \cite{sharma2018multiple} presented a dataset that comprises 8.3k episodes for a dual-arm robot that features tasks such as pushing with two hands.
Brohan \textit{et al.} \cite{brohan2022rt} showed that using a dataset containing 130k language-instructed episodes and various tasks enabled generalization to various backgrounds and objects in object manipulation, and focused mainly on grasping\footnote{In this research, \textit{pick} refers to the picking up of thin objects from a table; hence, \textit{grasp} is used to represent what is conventionally referred to as \textit{pick}.}, placing, and moving objects. In a subsequent study, Brohan \textit{et al.} \cite{rt22023arxiv} further demonstrated that the use of a large vision-language model enabled the emergence of task reasoning for unseen objects. 
In another study, a single-arm robot was used \cite{walke2023bridgedata}, and data from 60.1k episodes across 24 environments with open vocabulary were collected and trained, which encompassed 13 skills in various scenes, such as stacking and folding clothes.
Fang \textit{et al.} \cite{fang2023rh20t} created a multi-robot/multi-modal dataset containing 110k episodes of contact-rich, single-arm tasks, such as cutting or plugging, accompanied by language descriptions.
To the best of our knowledge, publicly available robot manipulation datasets at scale do not yet include examples of dual-arm fine manipulation (refer to Table \ref{tab:dataset_summary}), such as the grasping of sub-centimeter-level objects, picking up thin objects, or opening zippers placed on a table, which are difficult to grasp because of their small size. Our DAA dataset, accompanied by the relatively large 224k episodes,\footnote{Each episode is structured to encompass only one semantic task. For instance, in a robot's grasp-and-place action, ``grasp'’ and ``place'’ are considered to be separate episodes. As a result, the number of episodes cannot be directly compared with those in other datasets, such as RT-1.} fills this gap by featuring dual-arm, fine manipulation skills, and incorporating visual attention signals and dual-action labels to help learning. 

\section{Robot Framework}\label{sec:robot_framework}
\begin{figure}
    \centering
        \includegraphics[width=0.9\linewidth]{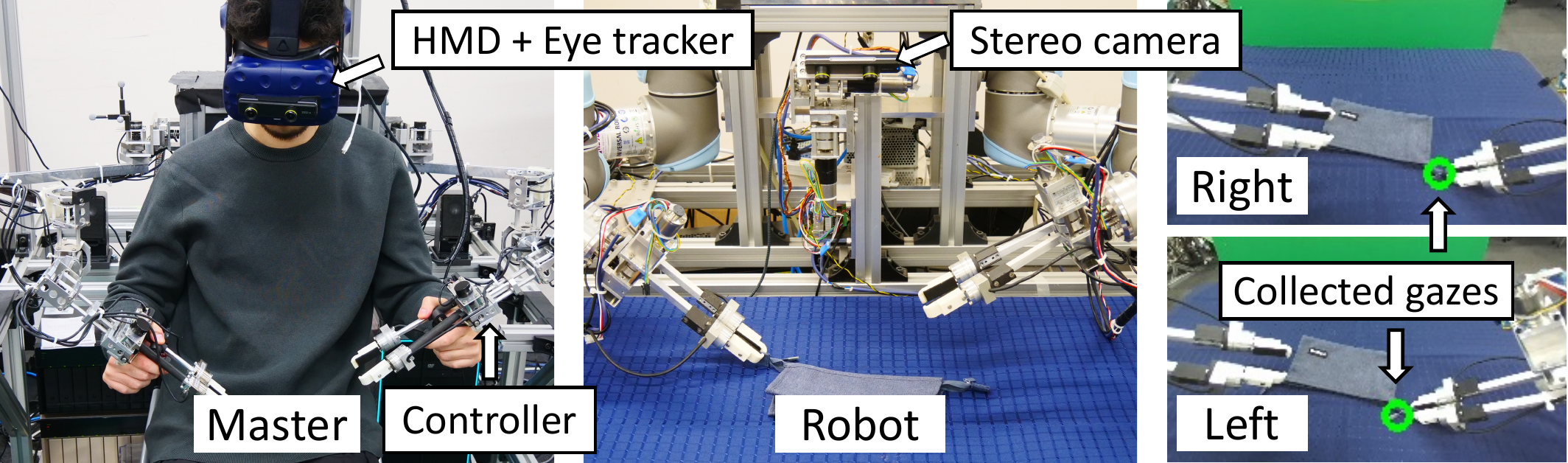}
        \caption{Robot and teleoperation framework with an eye tracker.}
        \label{fig:framework}
\end{figure}

In this study, we use the robot framework used in, for example, \cite{kim2020using,kim2021transformer}, to generate data for multi-task manipulation (Fig. \ref{fig:framework}). The robot used in this study consists of two UR5 robots equipped with custom-designed grippers. Human experts manipulate the robot using a master controller that mimics the UR5's Denavit–Hartenberg (DH) parameters. This configuration of the master controller has the advantage of naturally enabling humans to generate policies that consider the robot's embodiment. For example, policies can be generated within the range of solvable inverse kinematics \cite{kim2022training}. The robot's stereo camera (ZED-mini, Steterolabs) vision is observed through a head-mounted display (HMD, Vive Pro Eye, HTC) for intuitive teleoperation. An eye tracker inside the HMD measures the gaze coordinates of both eyes on the screen during the demonstration. 

\section{Dual-Action and Attention}
DAA is based on the architecture proposed in \cite{kim2022robot}. In this section, we explain each element of the architecture and then clarify the modifications made in the present study.

\subsection{Attention-Based Imitation Learning} \label{subsection:attention}
In the context of robot manipulation, much of the information input from the robot's sensors, such as background pixels, is irrelevant to the task. Visual and somatosensory attention mechanisms have been proposed as a method to minimize the covariate shift caused by such information. 

First, \textbf{visual attention} is an architecture inspired by the human retina structure and gaze that was designed to extract task-relevant information from a robot's vision. The human retina comprises a narrow central region of \textit{foveated vision} that receives high-resolution color information and \textit{peripheral vision} that captures low-resolution grayscale information\cite{Kirkwood1995simple}. During object manipulation, the human gaze strongly correlates with the object \cite{hayhoe2005eye} using high-resolution information from foveated vision for manipulation \cite{paillard1996fast}.

Our demonstration framework includes an eye tracker mounted on the HMD that measures the $x, y$ coordinates of where both of the human expert's eyes are looking on the screen. Based on this information, we train a gaze predictor to predict the gaze location in \textit{global vision} \footnote{This corresponds to human peripheral vision, but we have named it global vision because it also includes the central area (i.e., low-resolution foveated vision).} that a human is likely to look at while performing a task. The policy predictor receives foveated vision, that is, the high-resolution pixels around the gaze position predicted by the gaze predictor, as its vision input. Thus, the policy predictor can receive task-related object information included in foveated vision while effectively suppressing the covariate shift caused by changes in the background or other objects in global vision. Additionally, foveated vision allows for a significant reduction in the total number of pixels ($1280 \times 720 \rightarrow 256 \times 226$) while maintaining the high resolution in the pixel area necessary for the task, thereby enabling precise object manipulation skills. By contrast, in conventional robot learning (e.g., \cite{brohan2022rt}), the global vision is resized to a low resolution ($300 \times 300$). Using a high-resolution global vision to maintain detail will require excessive computational resources. Because of these advantages, the visual attention mechanism has been used in multi-object manipulation \cite{kim2020using,kim2022memory} and fine manipulation \cite{kim2021gaze,kim2022robot}.

Second, \textbf{somatosensory attention} in robotics was proposed for efficient information processing, particularly in dual-arm manipulation \cite{kim2021transformer} and manipulation with force/torque sensory information \cite{kim2022training}. This approach addresses covariate shifts caused by variations in a robot's limb positions between the training and testing phases. Unlike visual attention, which can be directed by the human gaze, somatosensory attention lacks a clear external signal. Therefore, in \cite{kim2021transformer}, the researchers used a Transformer-based self-attention mechanism to effectively process somatosensory inputs. In this research, the policy predictor's encoder corresponds to that mechanism.

\subsection{Dual-Action}
\begin{figure}
  \centering
  \vspace{0.0in}
  \begin{subfigure}[t]{.9\linewidth}
    \captionsetup{width=1.\linewidth}
    \captionsetup{justification=centering}
    \includegraphics[width=0.98\linewidth]{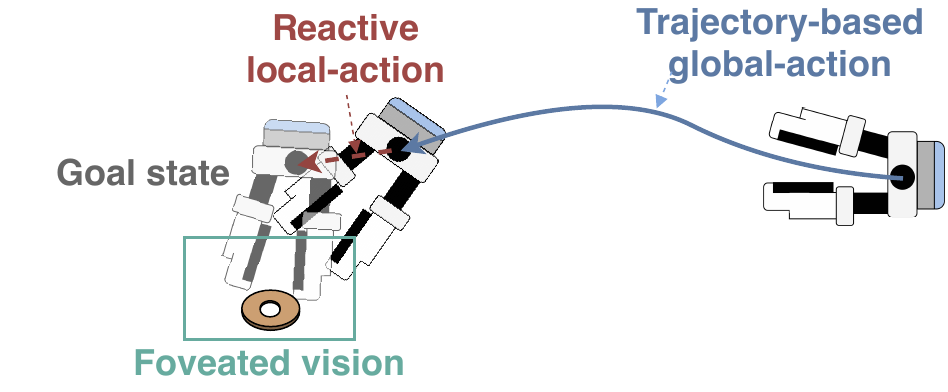}
    \caption{Dual-action separates the entire robot trajectory in both the demonstration set and during inference by the model into trajectory-based global-action and reactive local-action.}
    \label{fig:dual_action}
  \end{subfigure}%
  \\
  \begin{subfigure}[t]{.9\linewidth}
    \captionsetup{width=1.\linewidth}
    \captionsetup{justification=centering}
    \includegraphics[width=0.98\linewidth]{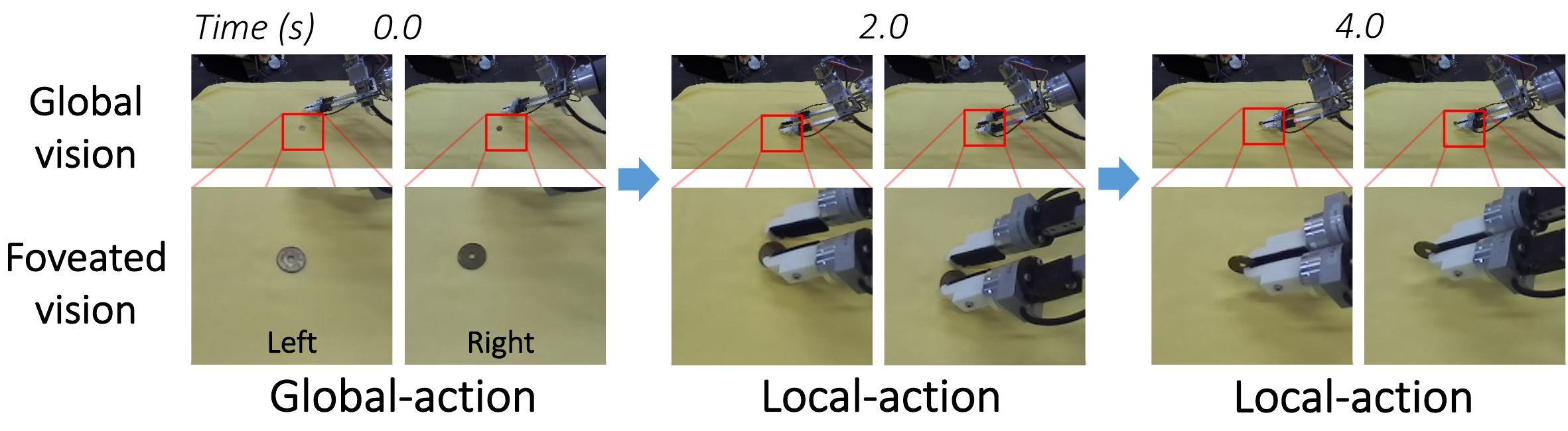}
    \caption{The gaze predictor uses low-resolution global vision to predict the gaze; hence, the policy predictor can use the high-resolution foveated vision of the target pixels for fine-grained manipulation. The action shifts from global to local when the end-effector is observed in foveated vision.}
    \label{fig:concept_fig}
  \end{subfigure}%
      \vskip\baselineskip
  \captionsetup{justification=centering}
  \caption{Brief explanation of dual-action and gaze-based attention.}
 \end{figure}

\textbf{Dual-action}, inspired by the human visuo-motor cognitive system \cite{paillard1996fast}, was proposed for precise object manipulation in \cite{kim2021gaze} and modified in \cite{kim2022robot}. This human system consists of two main parts: the \textit{movingness system} and \textit{displacement system}. The movingness system ballistically moves the hand rapidly around the target. By contrast, the displacement system uses information from foveated vision to achieve more detailed and accurate adjustments, thereby detecting positional errors. This dual system is fundamental to human visuomotor control and can be applied to robot object manipulation skill learning. In robot learning, it is divided into global-action and local-action, where each is responsible for fast movement that approaches the vicinity of the object and precise manipulation (Fig. \ref{fig:dual_action}). Specifically, in global-action, inspired by the ballistic movingness system, the trajectory from the current end-effector position to the starting point of local-action is predicted and executed. This is robust to the compounding error caused by the repetitive accumulation of small errors because the entire trajectory is predicted. Local-action is a reactive action that starts around the object and manipulates it. This allows for an immediate response to changes in the dynamics during object manipulation, which contributes to the success of delicate object manipulation. The combination of global and local-action results in robustness against compounding errors while achieving precise object manipulation. 

In this paper, we define the action shift from global-action to local-action as occurring when the end-effector is observed within foveated vision (Fig. \ref{fig:concept_fig}). This switching strategy is predicated on the assumption that the important interaction between the end-effector and the object, which is responsible for local-action, only occurs in foveated vision, which is selected by the gaze because the human's eye gaze is concentrated on the target object during manipulation \cite{hayhoe2005eye}. 

\section{Data Generation}\label{sec:multi_generation}

The dataset was generated by human demonstrators remotely operating a robot to perform various tasks.
A \textit{phase} is a cycle period from when the human teacher starts the remote operation to when they take a break. We repeated tens to hundreds of demonstration episodes per phase. Multiple tasks can exist in one phase. 
For example, in the phase of grasping nuts scattered on a table and placing them on a white plate, the instruction ``grasp nut with right white gripper’' and ``place nut on white plate with right white gripper'’ were repeated. In a single phase, background patterns (using tablecloths) and lighting conditions were changed multiple times. Foam floor tiles were placed on the surface of the table to prevent the robot from triggering emergency stops too frequently when it came into contact with the table, thereby enhancing safety during operation.

The dataset in this study includes, in addition to vision and robot action data, the following information. First, the \textbf{gaze} signal uses the eye tracker described in Section \ref{sec:robot_framework} to measure the gaze information of both eyes of the demonstrator. This information is important because it is highly correlated with the target object in the task. 
Second, the \textbf{dual-action label} in the dataset is designated as follows: a value of $0$ indicates a global-action, and a value of $1$ indicates a local-action. To simplify the annotation of dual-action for the entire dataset in this study, an automatic annotation method was adopted (Appendix \ref{appendix:automatic_annotation}). 
Finally, in this study, \textbf{language instructions} were used to provide clear directions for the tasks. Language instructions specify the target object and action, in addition to which robot arm to use: left, right, or both. 
Language instructions were annotated by humans for all tasks.

The dataset created in this study was divided into 11 groups according to the nature of the tasks. The task objectives and statistics of the dataset can be found in Table \ref{tab:large_scale_statistics}. The tasks targeted by this dataset are challenging. \textit{Grasp-and-place} mainly involves grasping small objects (e.g., bolts and candy) at the level of a few centimeters or sub-centimeters, where a few millimeters of error can lead to failure, which makes the grasping task difficult. In \textit{pick}, picking up thin objects, such as coins (Fig. \ref{fig:fig1_concept_dataset}), reduces the tolerance even further. In this study, the term \textit{pick} is specifically used to denote the action of lifting thin objects off a table. Therefore, the word \textit{grasp} is used to describe what is typically known as \textit{pick}. Opening or closing a pencil case also involves complex and delicate dual-arm manipulation, which takes into account the position and orientation of the small zipper tip during grasping. \textit{Needle-threading} and \textit{banana-peeling} involve the manipulation of deformable objects, and allow for very small manipulation tolerances \footnote{Banana-peeling and some data from the needle-threading tasks were taken from previous studies \cite{kim2021gaze,kim2022robot}.}.

\begin{table*}[hbt!]
\centering
\begin{tabular}{llccc}
\hlineB{2}
    \textbf{Task group}    & \textbf{Objective}              & \textbf{\# episodes}             & \textbf{Time (h)}    & \textbf{\# instructions}         \\
\hline \hline
Grasp-and-place      & Grasp and place various objects using one hand.            & 66,081                     & 41.8            &    302      \\
Pick                     & Pick up thin objects, such as coins and cards, using one hand.       & 10,936                     & 9.09        &       14        \\
Bottle             & Set up or knock down a bottle (inspired by \cite{brohan2022rt}).            & 8,114                      & 4.70          &       57      \\
Move-bowl     & Move bowls or plates with both hands to a target plate.               & 5,775                      & 3.33       &       48         \\
Pencil-case       & Open or close a pencil case using both hands.          & 23,650                     & 14.6       &       24        \\
Handkerchief     & Fold or unfold a handkerchief using both hands.                & 24,671                     & 13.3     &         155        \\
T-shirt            & Fold or unfold a t-shirt using both hands.             & 23,898                     & 15.1          &      356       \\
Lego             & Grasp Lego.            & 12,380                     & 9.60              &    75     \\
Miscellaneous      & Other tasks (e.g., rotating or stacking objects).            & 9,303                      & 6.88      &         52        \\
Needle-threading \cite{kim2021gaze}     & Pick up a thread and thread it through the needle.           & 9,975                      & 8.75            &    3       \\
Banana-peeling \cite{kim2022robot}      & Peel a banana.             & 29,427                     & 23.1    &       18           \\
\hline
\textbf{Sum}             &         & 224,210                    & 150      &        1,104       \\
\hlineB{2}
\end{tabular}
\caption{Task objectives and statistics.}\label{tab:large_scale_statistics}
\end{table*}

\section{Network Architecture}
\begin{figure*}[ht]
  \centering
  \includegraphics[width=0.8\linewidth]{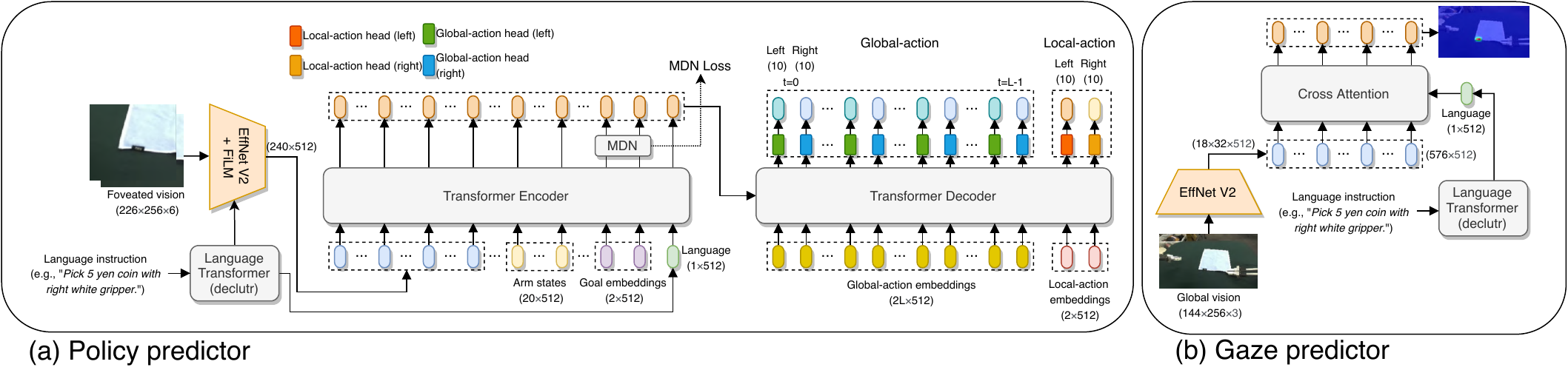}
  \caption{Neural network architectures. (a) The policy predictor outputs global-action and local-action through a Transformer encoder-decoder structure. It processes inputs such as high-resolution foveated vision attended by the gaze predictor, robot arm states, gaze coordinates, and language instructions. (b) The gaze predictor generates output by applying cross-attention to visual embeddings of global vision and language embeddings. The final gaze position is sampled based on the probability distribution of this output.}
  \label{fig:network}
\end{figure*}

The architecture of the policy predictor represents the evolution of the model proposed by \cite{kim2022robot} (Fig. \ref{fig:network} (a)). This neural network structure first processes foveated vision using EfficientNetV2, which is a parameter-efficient visual data processing model\cite{tan2021efficientnetv2}, then encodes the robot arm states and the predicted gaze position using a Transformer encoder. Subsequently, the Transformer decoder uses these encoded embeddings to output global-action and local-action. Global-action refers to a series of states up to a maximum of 20 time steps ahead, whereas local-action is the difference between the next state and the current state. Angles are represented in six dimensions for a continuous representation, as in \cite{zhou2019continuity}, rather than as Euler angles. Language instructions are first encoded using Deep Contrastive Learning for Unsupervised Textual Representations (\cite{giorgi2020declutr}), and then used as conditioned inputs for the Transformer encoder and FiLM \cite{perez2018film}-conditioned EfficientNet. FiLM was used, as in \cite{brohan2022rt}, to condition language-instructions to vision. However, unlike the study in \cite{brohan2022rt} that used global vision, in the present study, foveated vision is used; thus, FiLM-based conditioning cannot choose the appropriate arm to use. For instance, for a language input such as ``grasp nut with right white gripper,’’ there may not be any information about the hand in the foveated image, which prevents transferring information about the hand for use in downstream neural networks. Therefore, to condition hand information, language information was added as an additional embedding in the Transformer encoder. Additionally, the encoder predicts each task's goal position, which contributes to policy stabilization by producing manipulation policies based on goals \cite{kim2022robot}, which is trained using the mixture density network (MDN) loss \cite{bishop1994mixture} that fits the goal to the probability distribution represented by the Gaussian mixture model.

The main changes compared with the model in \cite{kim2022robot} can be summarized as follows: First, the model has been adapted to use language instructions so that, rather than having a specialized policy model for each task, a single policy model learns all tasks. Second, both global-action and local-action are inferred from the same neural network, and local-action also receives the robot's state and gaze coordinates. This is possible because of the use of the Transformer, which allows for attention to be directed to necessary somatosensory information, as shown in \cite{kim2021transformer}. Finally, goal position prediction was previously performed using a unimodal model with an $\ell_2$ loss, whereas DAA adopts a multimodal model using the MDN. This allows for the prediction of the probability distribution of goal positions, which enables the accurate prediction of target states, even in the presence of multiple objects.

The gaze predictor uses the cross-attention of the Transformer \cite{vaswani2017attention} (Fig. \ref{fig:network} (b)). Previous gaze prediction models (e.g., \cite{kim2020using}) predicted the gaze coordinates as a probability distribution using the MDN. 
However, training gaze prediction with the MDN was challenging under conditions involving multiple object scenes, which is a frequent occurrence in the training dataset. Therefore, in this study, left global vision is divided into an $18 \times 32$ grid, and the left gaze coordinate is transformed into a one-hot vector within this $18 \times 32$ grid of the left global vision, thereby solving it as a classification problem. Left global vision $144 \times 256$ serves as the input to EfficientNetV2, and the processed $18 \times 32$ visual features undergo cross-attention with language, as proposed in \cite{lynch2023interactive}, to condition language on vision. The output of cross-attention was trained using cross-entropy loss to predict the left gaze coordinate. Predicting both gazes independently (e.g., \cite{kim2022robot}) often led to predictions of different objects in multi-object environments. Therefore, in this study, the right gaze is calculated inversely using trigonometry. This is based on the predicted left gaze coordinates and depth information computed from the stereo camera.

\section{Experiment}\label{sec:multi_test}

We validated the generalization and robustness of multi-task trained DAA in a total of 7,815 trials for a real robot.

\subsection{Multi-Task Performance of DAA}\label{subsection:multitask_test}

We evaluated the generalization ability across various tasks and novel objects. To achieve this, 103 tasks with objects included in the training set and 39 tasks involving objects not included in the training set were attempted 9 to 36 times each. 
The test task setups are explained in Table \ref{tab:test_group_trials}. \textit{Pick} tests, which require picking up very thin objects, such as coins or plastic cards on a table, are differentiated from \textit{grasp} tests because the former tests demand distinct manipulation skills.
For the sake of experimental fairness, the background was uniformly set to green. The multi-task trained model was trained on the task groups \textit{grasp-and-place}, \textit{pick}, \textit{bottle}, \textit{move-bowl}, \textit{pencil-case}, \textit{handkerchief}, \textit{t-shirt}, and \textit{miscellaneous} from the dataset\footnote{\textit{Lego} was excluded because of its complexity, and \textit{needle-threading} and \textit{banana-peeling} were experimented on additionally in \ref{subsection:additional_exp}.}. Task-specific trained models were trained separately for each object in each task group.
In the testing of the proposed DAA, results for \textit{t-shirt} and \textit{miscellaneous} were excluded. The \textit{t-shirt} task failed because the t-shirt, which was large, occupied the edges of the camera vision, which led to significant errors in the eye tracker. Thus, the gaze predictor failed to predict the correct gaze coordinate, which resulted in the failure of the folding t-shirt task.

\begin{figure*}
  \centering
  \vspace{0.02in}
  \includegraphics[width=.98\linewidth]{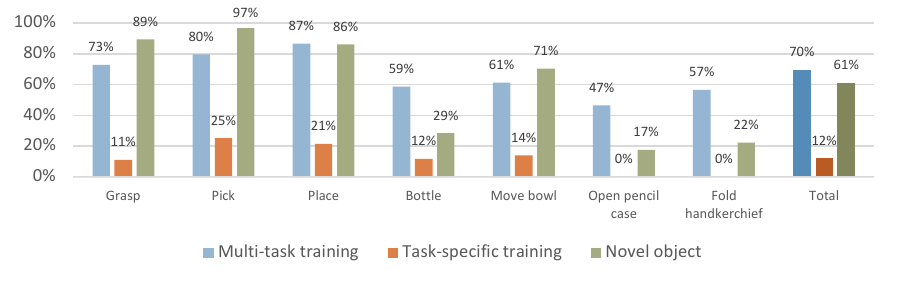}
  \captionsetup{width=.9\linewidth}
  \captionsetup{justification=centering}
  \caption{Manipulation results of multi-task and task-specific trained models.}
  \label{fig:multi_task_result}
\end{figure*}

\begin{table}[ht]
\centering
\begin{tabular}{llc}
\hlineB{2}
    \textbf{Test} & \textbf{Objective}                           & \textbf{\# of Trials} \\
\hline \hline
Grasp            & Grasp a target object.                       & 16-32              \\
Pick             & Lift a thin target object off a table.                     & 21                    \\
Place            & Place a bolt in a bowl or plate.      & 25                    \\
Bottle           & Knock down a bottle or set it \\
		& upright.          & 9                     \\
Move-bowl        & Support both sides of a bowl  \\ 
                 & and place it on a plate.              & 10                    \\
Open-pencil-case & Grasp the tip and zipper of \\ 
                 & a pencil case and open it.                                   & 30                    \\
Fold-handkerchief& Grasp the left and right parts of \\ 
                 & a handkerchief and fold it.         & 9                     \\
\hlineB{2}
\end{tabular}
\caption{Test setups.}\label{tab:test_group_trials}
\end{table}

The results for the multi-task trained model and task-specific trained model are shown in Fig. \ref{fig:multi_task_result}. The proposed multi-task model achieved a 69.6\% success rate across all tasks, whereas the task-specific model achieved a 12.2\% success rate, which indicates that multi-task learning significantly improved the success rate. Additionally, the multi-task model recorded a 61.5\% success rate on new objects, which suggests that the multi-task model is capable of generalizing its manipulation skills to new objects.

\subsection{Ablation Studies}

\begin{figure}
    \centering
    \begin{subfigure}[t]{0.9\linewidth}
        \includegraphics[width=\linewidth]{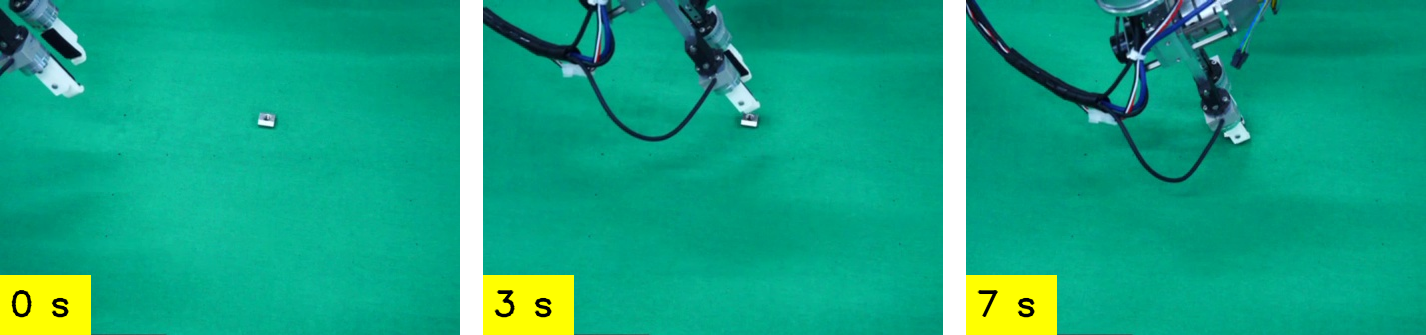}
        \caption{DAA (proposed) succeeded in grasping the target object.}
        \label{fig:proposed_concatenated}
    \end{subfigure}
    \\
    \begin{subfigure}[t]{0.9\linewidth}
        \includegraphics[width=\linewidth]{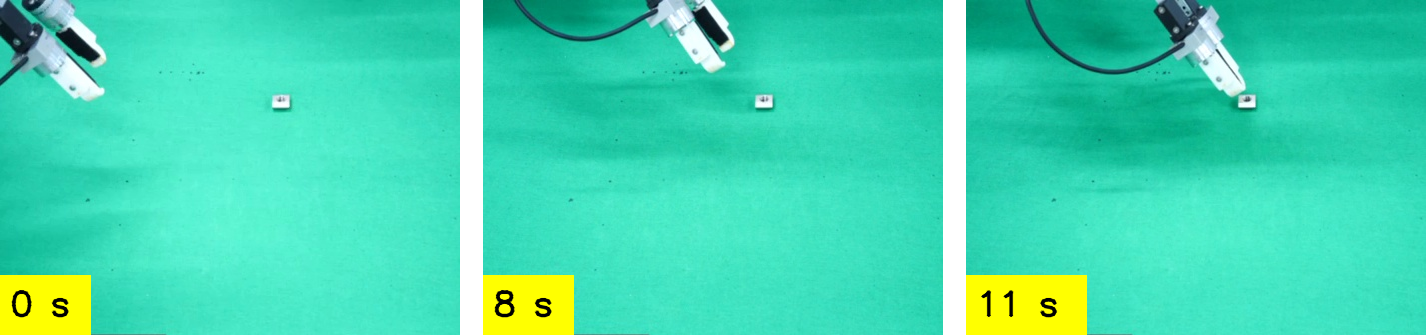}
        \caption{RT-1 (with gaze) on a smaller object failed to grasp the target object.}
        \label{fig:rt1_foveated_concatenated}
    \end{subfigure}
    \\
    \begin{subfigure}[t]{0.9\linewidth}
        \includegraphics[width=\linewidth]{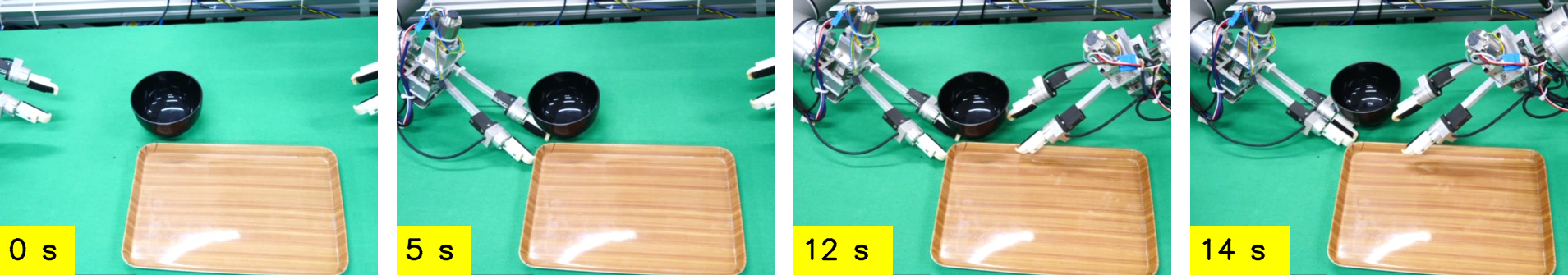}
        \caption{RT-1 (with gaze) on a relatively larger object failed to hold the target bowl.}
        \label{fig:rt1_foveated_black_bowl_concatenated}
    \end{subfigure}
    \caption{Comparison of DAA and RT-1. RT-1 (with gaze) moved the end-effector close to the target object, but it failed to achieve successful manipulation.}
    \label{fig:comparison_gada_rt1}
\end{figure}

\begin{figure*}
  \centering
  \vspace{0.02in}
  \includegraphics[width=.98\linewidth]{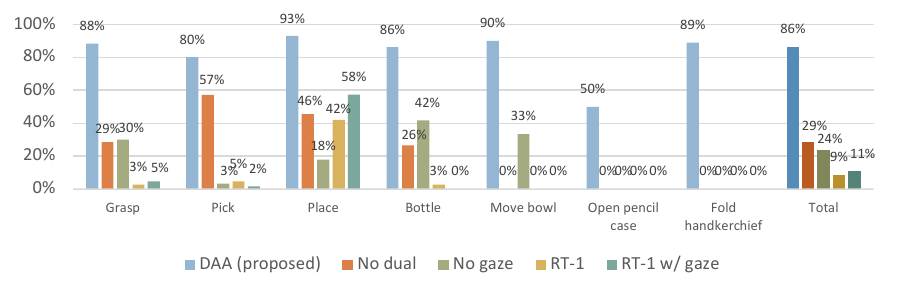}
  \captionsetup{width=.9\linewidth}
  \captionsetup{justification=centering}
  \caption{Results of the ablation studies.}
  \label{fig:multi_ablation}
\end{figure*}

DAA includes gaze-based visual attention and dual-action. To verify the importance of these elements in multi-task learning, in Fig. \ref{fig:multi_ablation}, DAA is compared across 26 tasks with a \textit{No gaze} model, which lacks the visual attention mechanism, but uses global vision instead of foveated vision, and a \textit{No dual} model, which lacks dual-action, and thus does not distinguish dual-action and considers all actions as reactive. All of these models have the same number of policy predictor parameters. Additionally, comparisons were made with RT-1 \cite{brohan2022rt}, and a modified version of RT-1 using foveated vision and somatosensory information (RT-1 with gaze, Appendix \ref{appendix:model_train}). 
As a result, both the \textit{No dual} and \textit{No gaze} models achieved lower success rates, which indicates the necessity of visual attention and dual-action. Furthermore, the RT-1s, as shown in Fig. \ref{fig:rt1_foveated_concatenated}, were not suitable for datasets that include tasks that require fine manipulation skills, although they achieved some success in \textit{place} tasks that did not require precision. This implies that the manipulation skill properties of our dataset are significantly different from those of RT-1, which is focused on generalizing to new objects or environments using a large dataset that mainly comprises pick-and-place tasks.

\subsection{Robustness Against Light Condition Variations}
\begin{figure}
    \centering
        \begin{subfigure}[t]{0.9\linewidth}
            \captionsetup{width=.8\linewidth}
            \captionsetup{justification=centering}
            \includegraphics[width=0.98\linewidth]{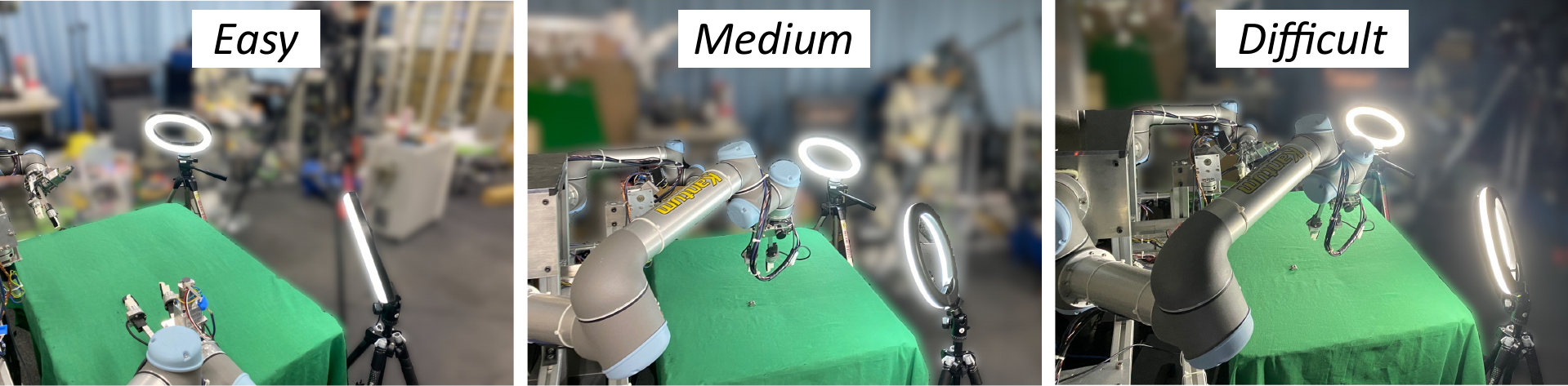}
            \caption{Different light conditions.}
            \label{fig:multi_light_conditions}
        \end{subfigure}%
        \\
        \begin{subfigure}[t]{0.99\linewidth}
            \captionsetup{width=.9\linewidth}
            \captionsetup{justification=centering}
            \includegraphics[width=0.98\linewidth]{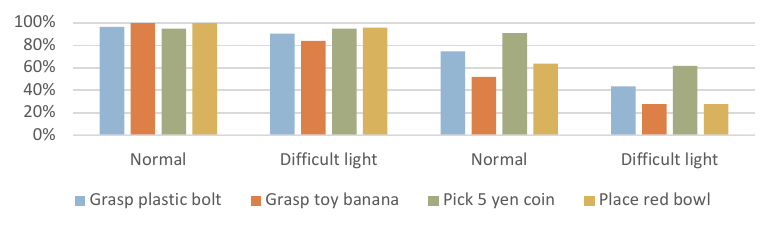}
            \caption{Comparison of multi-task and task-specific trained models for various light conditions.}
            \label{fig:multi_light_result_task_specific}
        \end{subfigure}%
        \caption{Test results for various light conditions.}
        \label{fig:multi_light_results}
\end{figure}

Changes in lighting conditions, such as luminance and shadow direction, alter various visual elements, and the DAA is required to adapt to these changes robustly. To verify the robustness of the DAA, the degree of lighting change was controlled as \textit{Easy}, \textit{Medium}, and \textit{Difficult} (Fig. \ref{fig:multi_light_conditions}), and the robot was tested on three grasping tasks. 
In Fig. \ref{fig:multi_light_result_task_specific}, the results of the proposed multi-task trained model and task-specific models are compared for four tasks. The results showed that the multi-task model recorded only a slight decrease in the success rate, even under \textit{Difficult} conditions, whereas the task-specific trained models experienced a significant drop in the success rate.

\subsection{Robustness Against Background Variations}

\begin{figure}
    \centering
        \begin{subfigure}[t]{0.99\linewidth}
            \captionsetup{width=.9\linewidth}
            \captionsetup{justification=centering}
            \includegraphics[width=0.98\linewidth]{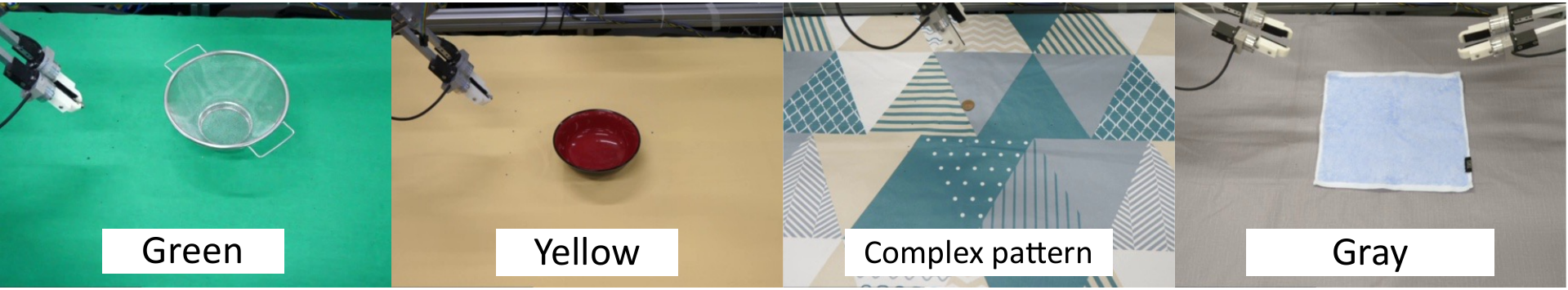}
            \caption{Various backgrounds.}
            \label{fig:multi_bgs}
        \end{subfigure}
        \\
        \begin{subfigure}[t]{0.99\linewidth}
            \captionsetup{width=.9\linewidth}
            \captionsetup{justification=centering}
            \includegraphics[width=0.98\linewidth]{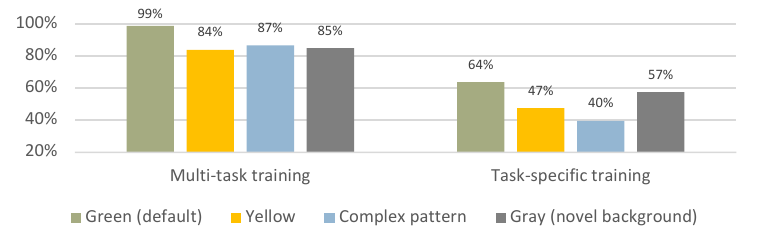}
            \caption{Comparison of multi-task and task-specific trained models on various backgrounds.}
            \label{fig:multi_bg_result_task_specific}
        \end{subfigure}%
        \caption{Test results for background changes.}
        \label{fig:multi_bg_results}
\end{figure}

The robustness of the multi-task trained DAA to changes in the robot's background was tested on four types of tablecloth backgrounds (Fig. \ref{fig:multi_bgs}): \textit{Green} (default background), \textit{Yellow} (another color included in the training dataset), \textit{Complex pattern} (a complex pattern included in the training data), and \textit{Gray} (a color not included in the training data).
In Fig. \ref{fig:multi_bg_result_task_specific}, the proposed multi-task trained model is compared with task-specific trained models on three tasks. These tasks were selected because the task-specific models achieved relatively high performance in them. As a result, it was observed that the success rate of the task-specific trained models, particularly in the case of complex patterned backgrounds, was significantly lower than in the case of the default background. By contrast, the multi-task trained model maintained a more consistent success rate across various backgrounds.

\subsection{Additional Results}\label{subsection:additional_exp}

\begin{figure}
    \centering
        \begin{subfigure}[t]{0.5\linewidth}
            \captionsetup{width=.9\linewidth}
            \captionsetup{justification=centering}
            \includegraphics[width=0.98\linewidth]{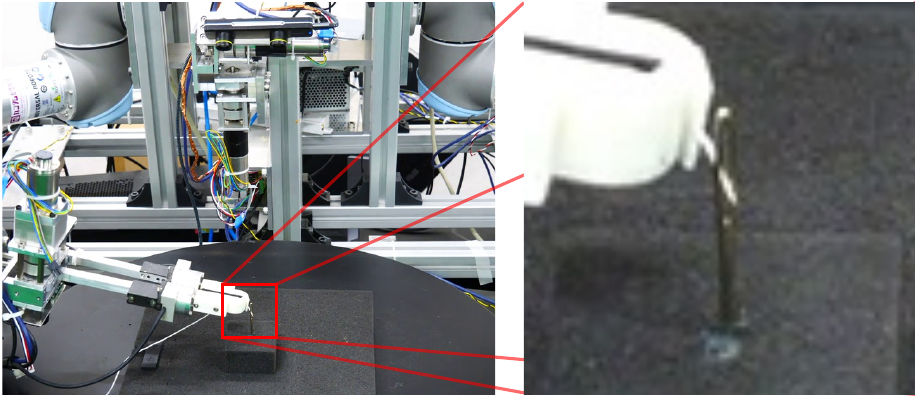}
            \caption{Needle-threading.}
            \label{fig:needle_threading}
        \end{subfigure}
        \begin{subfigure}[t]{0.3\linewidth}
            \captionsetup{width=.9\linewidth}
            \captionsetup{justification=centering}
            \includegraphics[width=0.98\linewidth]{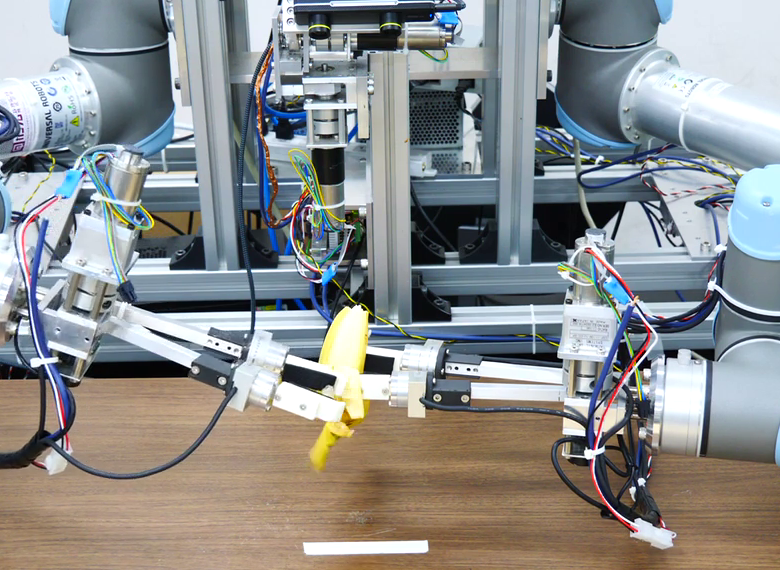}
            \caption{Banana-peeling.}
            \label{fig:banana_peeling}
        \end{subfigure}%
        \caption{DAA executes challenging manipulation tasks.}
        \label{fig:needle_banana}
\end{figure}

DAA was trained on challenging manipulation tasks (Fig. \ref{fig:needle_banana}) that were previously presented in \cite{kim2021gaze} and \cite{kim2022robot}. First, Table \ref{tab:needle_threading} shows that, for needle-threading, DAA achieved manipulation accuracy comparable with the results reported in \cite{kim2021gaze} \footnote{These results used different amounts of data (8.75 hours in the current dataset vs. 3.60 hours in \cite{kim2021gaze}). Because the environment including the lighting condition changed from \cite{kim2021gaze}, these results cannot be directly compared.}. It is noteworthy that, compared with a model trained from scratch, there was no performance improvement in the model that started training from the multi-task trained model in Section \ref{sec:multi_test}. This result suggests that multi-task training may not be effective for learning completely novel skills.

\begin{table}[hbt!]
\centering
\begin{tabular}{ccc}
\hlineB{2}
Previous work \cite{kim2021gaze} & DAA (scratch) & DAA (pretrained) \\ 
\hline \hline
81.25\% & 85.71\% & 85.19\% \\
\hlineB{2}
\end{tabular}
\caption{Needle-threading results.}
\label{tab:needle_threading}
\end{table}

The result of DAA for the banana-peeling task is  presented in Table \ref{tab:banana_peeling}, which demonstrates performance comparable with previous the study \cite{kim2022robot}. Because there was no performance improvement with DAA (pretrained) in the needle-threading task, only DAA (scratch) was tested.

\begin{table}[hbt!]
\centering
\begin{tabular}{cc}
\hlineB{2}
Previous work \cite{kim2022robot} & DAA (scratch) \\ 
\hline \hline
0.870 & 0.900 \\
\hlineB{2}
\end{tabular}
\caption{Banana-peeling results.}
\label{tab:banana_peeling}
\end{table}

\section{Conclusions and Discussion}\label{sec:multi_summary}

We addressed the challenge of imitation learning for dual-arm fine manipulation using a multi-task, at-scale dataset and DAA. We created and released a robot demonstration dataset containing 224k episodes for imitation learning, and implemented a multi-task learning agent by applying the dataset to DAA. This dataset is novel in that it includes visual attention gaze signals and dual-action labels, and it includes fine manipulation skills with deformable objects and/or with dual arms. We successfully extended DAA for learning dual-arm fine manipulation to multi-task learning using this dataset and observed an improvement in manipulation performance through multi-task learning.
Furthermore, the effectiveness of gaze-based attention was confirmed for fine object manipulation in multi-task learning. This architecture selectively extracts high-resolution pixels related to the target object in a scene. Additionally, the dual-action approach, which distinguishes between ballistic global-action for approaching the vicinity of the object and reactive local-action for precise manipulation, was also important.

The limitations of this study include, first, the use of only one type of robot framework. Considering that in studies such as \cite{open_x_embodiment_rt_x_2023}, researchers produced learning data for various types of robots, our framework's adaptability to novel robot frameworks can be regarded as relatively lower. However, we believe that the usability of our robot data in the robot research community can be maximized for the following reasons. First, by making the design of the robot public, we have enabled reproducibility. Second, we expect that the gaze-based attention mechanism, which removes the robot arm's image from policy inference and only includes the end-effector's visual appearance, will enhance usability because the end-effector is relatively easy to replicate. The second limitation is that, in this study, we focused on the generalization of fine dual-arm manipulation skills and did not address the acquisition of semantic reasoning. This would require a combination of  DAA dataset with large vision-language models, as discussed in \cite{rt22023arxiv}; however, exploring the emergence of such features for fine manipulation tasks is challenging and may require large language-action pairs and computational power.
Finally, we found that acquiring fine manipulation skills remains challenging and demands a substantial amount of data. 
We are optimistic that integrating spatial models in robotics, such as SE(3)-equivariance (e.g., \cite{ryu2022equivariant}), into the learning process of our model may improve sample efficiency through the acquisition of object pose-invariant manipulation skills.

\appendix

\section{Appendix}

\subsection{Automated Dual-Action Annotation} \label{appendix:automatic_annotation}

For dual-action, human annotators manually annotated approximately 6,000 episodes. Based on this data, a neural network that considers time series was trained with a three-layer Transformer encoder following EfficientNetV2 to automatically annotate the remaining data. The trained model was also used to predict dual-action labels during the test.

\subsection{Model Details} \label{appendix:model_train}

The DAA's architecture comprises 12 layers each for the encoder and decoder, which have an embedding size of 768 and feedforward network dimension of 3072 within the Transformer layers. The total number of parameters is 350M. 
The model was trained on a DGX-1 (v100 $\times 8$) for 30 epochs over approximately 5 days. The dataset underwent a 95/5 training/validation split. The first 10 epochs involved a learning rate warmup, where the learning rate was increased from $1e-6$ to $1e-5$, followed by a gradual decrease in the learning rate over 20 epochs using cosine decay \cite{loshchilov2016sgdr}.
Training was conducted with a batch size of 8.

The training of RT-1 was based on \url{https://github.com/lucidrains/robotic-transformer-pytorch}. Additionally, the model was modified to use images at the current time step rather than accumulating six images as proposed in \cite{brohan2022rt} because accumulation did not work in our environment and to reduce the training time, which was roughly half the speed of DAA, even without accumulation. The number of parameters of the RT-1s was 198M. Global vision to $224 \times 224$ and an output dimension of $10+10$ were used to represent dual-arm actions. RT-1 with gaze inputs dual-arm states and gaze coordinates into Conv as additional channel data, in addition to the resized foveated vision ($224\times224$). This structure was designed to avoid making too many changes to the original RT-1 structure.

\bibliographystyle{IEEEtran}
\bibliography{IEEEfull}
\end{document}